\pdfoutput=1

\documentclass[11pt]{article}

\usepackage[final]{acl}

\usepackage{times}
\usepackage{latexsym}

\usepackage[T1]{fontenc}

\usepackage[utf8]{inputenc}

\usepackage{microtype}

\usepackage{inconsolata}

\usepackage{graphicx}
\usepackage{amsmath}
\usepackage{bbm}
\usepackage{multirow}
\usepackage{pifont}
\usepackage{amssymb}
\usepackage{booktabs}
\usepackage{enumitem}
\usepackage{microtype}
\usepackage{array}

\usepackage{times}
\usepackage{latexsym}
\usepackage{amssymb}
\usepackage{tabularx}
\usepackage{natbib}
\usepackage{makecell}
\usepackage{url}
\title{ReflectDiffu: Reflect between Emotion-intent Contagion and Mimicry for Empathetic Response Generation via a RL-Diffusion Framework}

\author{
 \textbf{Jiahao Yuan\textsuperscript{1,2}\thanks{jhyuan.cs@gmail.com}},
 \textbf{Zixiang Di\textsuperscript{2}},
 \textbf{Zhiqing Cui\textsuperscript{3}},
 \textbf{Guisong Yang\textsuperscript{1}\thanks{Corresponding author: gsyang@usst.edu.cn}},
 \textbf{Usman Naseem\textsuperscript{4}}
\\
 \textsuperscript{1}University of Shanghai for Science and Technology \\
 \textsuperscript{2}East China Normal University
 \\
 \textsuperscript{3}Nanjing University of Information Science and Technology
 \\
 \textsuperscript{4}Macquarie University
\\
}

\begin{document}
\maketitle
\begin{abstract}
Empathetic response generation necessitates the integration of emotional and intentional dynamics to foster meaningful interactions. Existing research either neglects the intricate interplay between emotion and intent, leading to suboptimal controllability of empathy, or resorts to large language models (LLMs), which incur significant computational overhead. In this paper, we introduce ReflectDiffu, a lightweight and comprehensive framework for empathetic response generation. This framework incorporates emotion contagion to augment emotional expressiveness and employs an emotion-reasoning mask to pinpoint critical emotional elements. Additionally, it integrates intent mimicry within reinforcement learning for refinement during diffusion. By harnessing an \textit{intent twice} reflect mechanism of \textit{Exploring-Sampling-Correcting}, ReflectDiffu adeptly translates emotional decision-making into precise intent actions, thereby addressing empathetic response misalignments stemming from emotional misrecognition. Through reflection, the framework maps emotional states to intents, markedly enhancing both response empathy and flexibility. Comprehensive experiments reveal that ReflectDiffu outperforms existing models regarding relevance, controllability, and informativeness, achieving state-of-the-art results in both automatic and human evaluations.
\end{abstract}

\begin{figure}[ht]
  \includegraphics[width=0.99\linewidth]{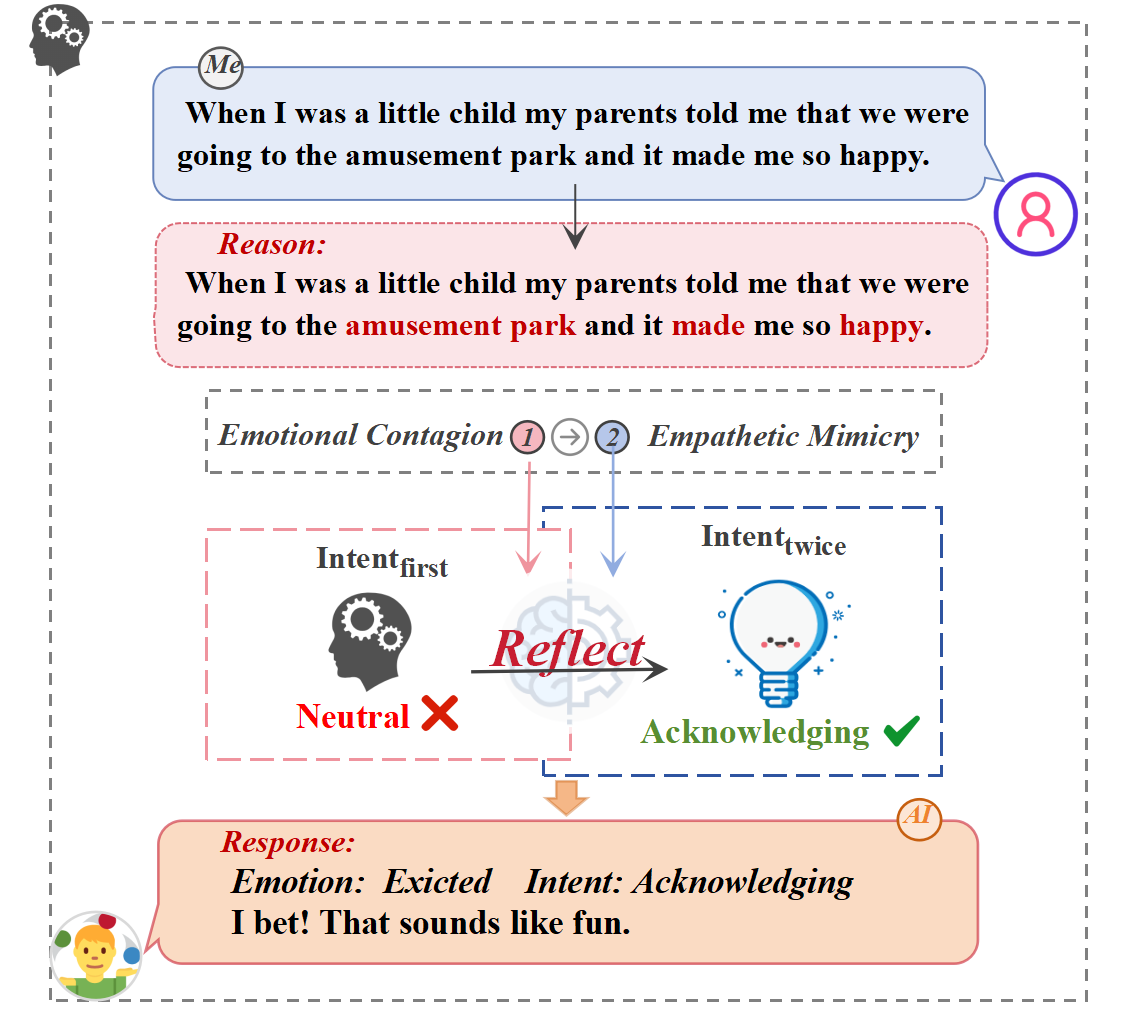}
  \caption{An example from the \texttt{EMPATHETICDIALOGUES} dataset involves incorporating Emotional Contagion and Mimicking, using \textit{intent twice} mechanism to enhance empathy.}
  \label{fig:intro}
\end{figure}

\section{Introduction}
Empathetic dialogue generation endows dialogue models with human-like emotional capabilities to recognize, understand, and express emotions \cite{davis1990empathy,cuff2016empathy}. In psychology, empathy mechanisms are empirically linked to sociological studies on emotional contagion \cite{hatfield1993emotional} and empathetic mimicry \cite{carr2003neural}. Recent research has delved into various aspects of empathetic mechanisms in chatbots, including dynamically tailoring responses based on perceived emotional triggers \cite{gao2021improving,gao2023cab} or mimicking empathetic emotions \cite{majumder2020mime,bi2023diffusemp}. 

Existing models typically generate responses based on either mimicking emotional states \cite{lin2019moel,majumder2020mime} or incorporating external knowledge including multi-resolution strategies \cite{li2020empdg}, commonsense reasoning through predefined sources \cite{li2022knowledge} or extracted via COMET \cite{hwang2021comet, sabour2022cem}, and multi-grained signals including causes \cite{bi2023diffusemp,hamad2024asem} to enhance contextual understanding. 

Recent advances in large language models (LLMs) \cite{dubey2024llama,yang2024qwen2} have promoted several empathetic dialogue models utilizing multiple-stage Chain-of-Thought (CoT) \cite{chen2024cause,hu2024aptness} with fine-tuning \cite{zhang2020dialogpt,cai2024empcrl}. However, their unstable performance \cite{lu2022fantastically,xie2023ask,yuan2024cultural} and reliance on external knowledge and high training costs \cite{kaplan2020scaling,yuan2024cultural} complicate practical implementation. Consequently, current research focuses on enhancing small-scale empathetic models through empathy mechanisms \cite{majumder2020mime,gao2023cab,zhou2023case,wang2024ctsm} as a more lightweight, practical alternative to LLMs. In summary, lightweight empathetic models encounter three major limitations: (1) They primarily rely on supplementary knowledge signals~\cite{gao2023cab} rather than underlying psychological mechanisms, which impedes controllability and empathetic capability. (2) They often overlook the internal mechanisms behind emotional causes, emotions, and intents, which rely heavily on external knowledge or pre-trained annotators \cite{bi2023diffusemp,chen2024cause}, resulting in hard-coded enhancements rather than genuine understanding and iterative correction, thus impacting empathy, diversity and flexibility. (3) There is a shortage of multi-task datasets for emotion reason masking, intent prediction, and empathetic dialogue. Most models rely on supplementary datasets for auxiliary tasks \cite{li2024reinforcement, bi2023diffusemp}, which does not guarantee that the advantages of multi-task training are fully realized or effectively aligned.

To address above limitations, we propose ReflectDiffu, a lightweight and comprehensive framework for empathetic response that seamlessly blends emotional contagion with intent prediction through a reflect mechanism. In sociology, empathetic actions are caused by emotional contagion \cite{hatfield1993emotional} and empathetic mimicking \cite{de2017mammalian}, which indicate an imitation feedback mechanism between human emotions and intentional actions \cite{rizzolatti2005mirror,iacoboni2009imitation}, as depicted in Figure~\ref{fig:intro}. Our key contributions include:
\begin{itemize}[itemsep=0.1em, parsep=0pt, topsep=0pt]
    \item We introduce a novel empathetic framework, ReflectDiffu, guided by sociological theories on emotional contagion and empathetic mimicry to improve empathy.
    \item We propose an \textit{intent twice} mechanism, termed \textit{Exploring-Sampling-Correcting} guided by reflect mechanism to align emotion with intent and minimize empathetic response misalignment caused by emotional misrecognition. 
    \item We conducted extensive experiments demonstrating that ReflectDiffu outperforms state-of-the-art models in both automatic and human evaluations.
\end{itemize}

\section{Related Work}

\subsection{Empathetic Response Generation}

Empathetic response generation entails recognizing emotional states and producing suitable emotional responses \cite{davis1983measuring, rashkin2018empathetic}. Early studies primarily aimed at generating emotion-specific responses based solely on emotional states \cite{lin2019moel, majumder2020mime}, but faced challenges with the explainability and controllability of empathy. Additionally, reinforcement learning (RL) has been employed to refine dialogue policies, with works like 
 \cite{li2024reinforcement} leveraging policy-based RL to optimize empathetic response generation. Recent studies have integrated external commonsense reasoning \cite{li2022knowledge, zhou2023case}, predefined knowledge \cite{li2020empdg, gao2023cab, wang2024ctsm} or pre-trained causal factors \cite{hwang2021comet, sabour2022cem} to enhance emotional perception, but they overlook the established interconnections among factors \cite{de2017mammalian}, which restricts deeper interpretability and empathy. 

Unlike previous approaches, ReflectDiffu introduces reflection interconnection to systematically convert emotional dimensions into actionable intents, thereby improving empathy.
\subsection{Generative Model for Dialogue Generation}
Generative models have exhibited outstanding performance, facilitating text generation. \citet{majumder2020mime} pioneered introducing Variational Autoencoders (VAEs) \cite{park2018hierarchical} to imbue text with empathetic expressions. Further research by \citet{gao2023cab} introduced latent variables \cite{sohn2015learning} accounting for cognition, affection, and behavior to better model emotional dependencies in dialogues. 

Subsequently, Denoising Diffusion Probabilistic Models (DDPMs) \cite{ho2020denoising} stands out for generating high-quality samples via iterative denoising \cite{li2025expensive}. \citet{hoogeboom2021argmax} and \citet{austin2021structured} paved the way for character-level text generation with diffusion. \citet{li2022diffusion} uses an embedding and rounding strategy and additional classifiers for controllable text generation. \citet{gong2022diffuseq} introduces a classifier-free diffusion model for dialogue generation. \citet{bi2023diffusemp} incorporated multi-grained control signals but their multi-stage pre-training approaches increase computational costs and the difficulty of practical implementation.

As far as we are aware, we are among the first to achieve multi-task empathetic response generation using reinforcement learning within diffusion guided by psychological knowledge.

\begin{figure*}[ht]
  \includegraphics[width=0.99\linewidth]{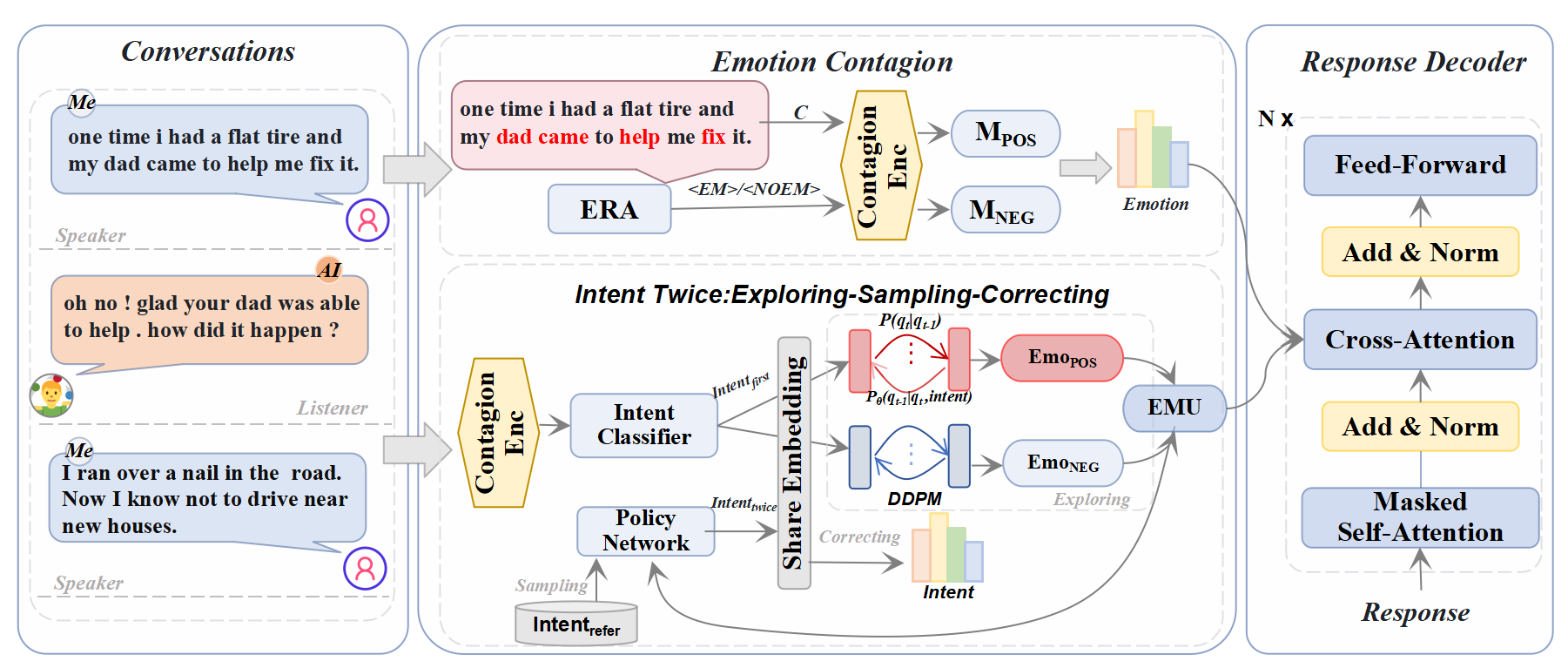}
  \caption{Architecture of our model (ReflectDiffu), which comprises three primary components: Empathetic Imitation influenced by Emotional Contagion, \textit{Intent Twice: Exploring-Sampling-Correcting} Mechanisms and a Response Decoder. }
  \label{fig:model}
\end{figure*}
\vspace{-0.3em}
\section{Methodology}
Our model, ReflectDiffu, is inspired by sociological studies on emotional contagion \cite{hatfield1993emotional} and empathetic mirroring \cite{de2017mammalian}, which suggest that empathy involves aligning emotional states and mimicking empathetic behavior in interpersonal interactions. Positive emotions are met with positivity, while in situations involving negative emotions, the empathetic response strategy incorporates a congruent emotional tone infused with positivity and a precise empathy intent to resonate deeply with speakers \cite{majumder2020mime,chen2022emphi}.

ReflectDiffu comprises two essential components: an \textit{Emotion-Contagion Encoder}, enhanced with an emotional reasoning annotator for improved semantic comprehension, and a \textit{Rational Response Generation Decoder} guided by an \textit{Intent Exploring-Sampling-Correcting} mechanism, which mirrors human reflective dialogue behavior to enhance empathy with robustness. The architecture of our model is depicted in Figure~\ref{fig:model}.

\subsection{Task Definition}
The conversation history consists of multiple interactions between a user and a chatbot, represented as $C = [c_0, c_1, \dots, c_{n-1}]$, where $n$ denotes the number of conversation rounds. Each utterance, $c_i$, is tokenized into a sequence of words:$C = [w_0^0, w_1^0, \dots, w_0^1, w_1^1, \dots, w^{n-1}_{m-1}]$. The primary aim is twofold: accurately discerning users' emotional state, denoted as $emo$, and formulating an empathetic response, $c_n$. Additionally, We introduce two auxiliary tasks: emotion reasoning annotation and intent prediction. Within $c_i$, emotional keywords are marked with the tag \texttt{<$em$>}, while non-emotional words are labeled \texttt{<$noem$>}. The chatbot also predicts the underlying conversation intent, $intent$, based on the entire dialogue sequence $C$.

\subsection{Multi-task Emotion-Contagion Encoder}
\paragraph{Emotion Reason Annotator.}
The Emotion Reason Annotator (\textit{ERA}) identifies emotional cues and generates reasoning masks within conversational turns. To efficiently handle labeled data for downstream NER tasks, we follow \citet{bogdanov2024nuner} to utilize a LLM for data annotation and conduct distillation training with other models. \textit{ERA} builds upon BERT \cite{devlin2019bert}, an attention-based semantic composition network, and conditional random fields (CRF) to effectively annotate emotional phrases with tags in the sequence $r$ consisting of \texttt{<$em$>} or \texttt{<$noem$>} and reasoning representations $\tilde{h}$ described in Appendix~\ref{subsec:appendix-era}.

\paragraph{Emotion-Contagion Encoder.}
The Emotion-Contagion Encoder incorporates the reasoning masks learned by \textit{ERA} into a transformer encoder to emulate emotional contagion.

Given that the reasoning masks $r$ for \texttt{<$em$>} or \texttt{<$noem$>} are only applied to users’ emotional state, the chatbot's $r$ is always set as \texttt{<$noem$>} because empathy is user-oriented, the distinction between users’ states and system states has been made. Therefore, unlike previous methods\cite{sabour2022cem,gao2023cab,bao2024multi}, we enhanced the context embedding by defining it as the sum of three embeddings ($E^C$): semantic word embedding ($E^W$), positional embedding($E^P$), and reason embedding ($E^R$), incorporating the \texttt{<$em/noem$>} into the final embedding, $E^C$, formally:
\begin{equation}
\label{eq:embed}
E^C = E^W(C) + E^P(C) + E^R(C),
\end{equation}
where $E^W(C), E^P(C), E^R(C) \in \mathbbm{R}^{D_{emb}}$.

Then, each token \( w_i^j \) in $E^C$  is transformed into its vector representation utilizing the context embedding $E_C$. Following the existing methods \cite{gao2023cab,wang2024ctsm,hamad2024asem}, we use a transformer encoder with one additional token, $CTX$, prepended to gain the speaker context. The transformer encoder, denoted as $\text{TRS}_\text{Enc}$, encodes the flattened $E^C$ into a context representation $H$:
\begin{equation}
\label{eq:trs}
H=\texttt{TRS}_\texttt{Enc}(E^C(C)).
\end{equation}
Finally, given the token-level context representation $H$ and the reasoning representations $\tilde{h}$ obtained from \textit{ERA}, we enhance $r$ by integrating it with $\tilde{h}$ through an attention layer and meaning aggregation, yielding overall representation $Q$, formally:
\begin{equation}
\label{eq:agg}
Q = \texttt{mean-pooling}(\texttt{Attention}(H, \tilde{h})).
\end{equation}

\paragraph{Contrastive-Experts Emotion Classification.}\label{subsubsec:emo}
Inspired by \cite{lin2019moel,majumder2020mime,chen2022hierarchical}, we put forward two-expert models \textit{C-Experts}: $\mathbbm{M}{pos}$ for positive emotions and $\mathbbm{M}{neg}$ for negative emotions to enhance emotion recognization by exploiting each model's proficiency. Neutral emotions are addressed via a voting mechanism between experts, yielding the candidate emotion probability distribution $p$ as follows:

\begin{small}
\begin{align}
\label{eq:s}
p = 
\begin{cases}
\texttt{softmax}(W_{\textit{neg}} \mathbf{E_{\textit{emo}}} Q) & \textit{if } v = \textit{n}_{\textit{neg}} \\
\texttt{softmax}(W_{\textit{pos}} \mathbf{E_{\textit{emo}}} Q) & \textit{if } v = \textit{n}_{\textit{pos}} \\
\texttt{Voting}(W_{\textit{neg}} \mathbf{E_{\textit{emo}}} Q),W_{\textit{pos}} \mathbf{E_{\textit{emo}}} Q))) & \text{if } v = \textit{n}_{\textit{neu}} \\
\end{cases}
\end{align}
\end{small}
where $v$ represents the maximum count among $\text{n}_{\text{neu}},\text{n}_{\text{pos}},\text{n}_{\text{neg}}$ within a batch, based on preliminary real-time sentiment analysis via VADER\cite{hutto2014vader}. $W_{\textit{pos}}$ and  $W_{\textit{neg}}$ are trainable weight matrices. $\texttt{Voting(*)}$ is a soft voting mechanism based on experts $\mathbbm{M}{pos}$ and $\mathbbm{M}{neg}$.

Additionally, we customize the $n_{\text{emo}}$ NT-Xent loss ($n_{\text{emo}}$ = 32), denoted as $L_{\text{NTX}}$, using pseudo labels to enhance context representation learning\cite{chen2020simple,zheng2023robust} on $Q$, while utilizing cross-entropy loss, $L_{\text{ce}}$, for classification. Consequently, the overall loss for emotion classification, $L_{\text{em}}$, is detailed in Appendix~\ref{subsec:appendix-loss}.
\subsection{Multi-task Rational Response Generation Decoder}
Building on psychological works\cite{hatfield1993emotional,de2017mammalian}, we propose that empathy involves mirroring users' emotions, responding positively to positive emotions and combining support with optimism for negative states. To enhance emotional encoding and empathetic expression with controllability, we conceptualize response intentions as actions\cite{chen2022emphi,gao2023cab}.
Unlike existing methods \cite{bi2023diffusemp,chen2022emphi}that rely on signals from externally fine-tuned classifiers, our multi-task response decoder integrates reinforcement learning into a diffusion framework\cite{ho2020denoising,gong2022diffuseq} to refine intent and enhance empathy, integrating Intent Twice, Emotion-Intent Mimicry, and Response Decoding to ensure coherent and empathetic interactions.

\subsubsection{Intent Twice: Exploring-Sampling-Correction.}
\label{subsubsec:intent_twice}
\paragraph{Exploring: First Intent Initialization.}
To enrich the contextual representation $Q$ with extra precise intent information, we consider both internal and external factors to score each candidate intent. In particular, we fine-tune a \texttt{BERT} classifier on the \texttt{EMPATHETICINTENT} dataset \cite{chen2022emphi} offline to obtain the intent distribution $\textit{p}_{\textit{intent}}$. Following a similar procedure as in Section~\ref{subsubsec:emo}, we compute $\textit{p}_{\textit{semantic}}$ online using similarity metrics and combine the two distributions to re-rank the intents so that we can get a more accurate first intent prediction $\textit{Intent}_{\textit{first}}$:
\begin{equation}
\label{eq:intentf}
\textit{Intent}_{\textit{first}} = \textit{p}_{\textit{semantic}} + \alpha \textit{p}_{\textit{intent}}.
\end{equation}
Here, $\alpha$ is a hyperparameter that balances internal and external factors.

\paragraph{Sampling: RL-Diffusion for Intent Twice.}
Inspired by \cite{majumder2020mime,de2017mammalian}, we hypothesize that empathetic behavior requires mimicking user emotions and integrating references to common emotion-corresponding actions with one’s own cognitive process when learning empathic expression\cite{majumder2020mime}. Hence, the alignment between current emotions and inferred intents with universal intents, denoted as $\textit{Intent}_{\textit{refer}}$, is crucial for refining intention predictions, especially when errors arise in expert emotion recognition. To enhance the accuracy of action predictions and improve both controllability and effectiveness of empathetic responses, we integrate policy-based reinforcement learning (RL) within Denoising Diffusion Probabilistic Models (DDPMs)\cite{ho2020denoising,gong2022diffuseq}, to sample more accurate and universal intents. Our framework leverages the exploration-exploitation trade-off $\mathbbm{M}_{p}$ to balance the learned intent actions and the sampling of new empathetic actions. When previous emotion recognition errors occur, \textit{intent twice} mechanism can alleviate emotional misrecognition and correct wrong intents by sampling universal intents.

To define $\textit{Intent}_{\textit{refer}}$ for reference, we perform a statistical survey for each emotion to find the top-$n$ empathetic intention actions. The optimal value of $n$ is 3 as shown in Table~\ref{tab:emo_act}. We provide an experimental analysis for $n=3$ in Appendix~\ref{subsec:Explanation_of_hyperparameter_n_of_intent_infer}.

\begin{table}[ht!]
    \centering
    \small
    \resizebox{\linewidth}{!}{
    \begin{tabular}{>{\centering\arraybackslash}p{0.55\linewidth}|>{\centering\arraybackslash}p{0.45\linewidth}}
        \hline
        \textbf{Emotion} & $\textbf{Intent}_\textbf{refer}$ \\        
        \hline
       surprised, proud, impressed, nostalgic, trusting, faithful, prepared & acknowledging, encouraging, neutral \\
        \hline
        excited, confident, joyful, grateful, content, caring, faithful & encouraging, sympathizing, acknowledging\\
        \hline
        angry, disappointed  & consoling, suggesting, encouraging\\
        \hline
        hopeful, sentimental & encouraging, wishing, consoling \\
        \hline
        anticipating, lonely, afraid, anxious, guilty, embarrassed, sad, apprehensive, terrified, jealous& consoling, encouraging, neutral \\
        \hline
        hopeful, sentimental & encouraging, wishing, consoling \\
        \hline
    \end{tabular}
    }
    \caption{Mapping of Emotion-Group to Top-3 Universal Intents for Reference}
    \label{tab:emo_act}
    \vspace{-1em}
\end{table}

\subparagraph{State Representation: Emotion Mimicry Unit.} Emotion Mimicry Unit(\textit{EMU}) initially splits the emotion-contagion encoding $Q$ into positive and negative polarity representations following emotion grouping \cite{majumder2020mime}, but with $\textit{Intent}_{\textit{first}}$ guidance in diffusion. We train two distinct DDPMs for positive-polarity representation $\textit{Emo}_\textit{pos}$ with $\textit{L}_{\textit{kl}_\textit{pos}}$ and negative-polarity representation $\textit{Emo}_\textit{neg}$ with $\textit{L}_{\textit{kl}_\textit{neg}}$ following emotion-group\cite{majumder2020mime}, we integrate the captured nuances of each emotional polarity with the content encoding $H$ to obtain state $\textit{Emo}_\textit{fused}$.

Given the emotion-contagion encoding $Q$ and a fixed step $t$, the diffusion process iteratively adds Gaussian noise $\epsilon \sim \mathcal{N}(0, I)$ to $Q$ over $t$ steps:
\begin{equation}
\small
Q_{t}=\sqrt{1-\beta_t} Q_{t-1} +\sqrt{\beta_t} \epsilon.
\end{equation}
Here, $Q_{t}$ denotes emotion-contagion encoding at time step $t$, and $\beta _t \in [1 \mathrm{e}-5, 5 \mathrm{e}-2]$ is the noise level at time step $t$.To recover the corrupted encodings $q_{t}$ to their original context representation, we propose an intent-aware Conditional Variational Auto Encoder(CVAE) $\mathcal{M}_\theta$ that predicts the noise $\epsilon$ at each step, motivated by \citet{park2018hierarchical,chung2022diffusion}:

\begin{small}
\begin{equation}
\tilde{Q}_{t - 1}=\frac{1}{\sqrt{1-\beta_t}}\left(Q_{t}-\frac{\beta_t\mathcal{M}_\theta(Q_{t},t,Intent_{\textit{first}})}{\sqrt{1-\sum_{s = 1}^{t}\beta_s}} \right).
\end{equation}
\end{small}
Here, $\tilde{Q}_{t - 1}$ represents the reconstructed encoding, and $\theta$ denotes the parameters of $\mathcal{M}_\theta$.
Finally, we integrate with context encoding $H$ via cross-attention to get state $\textit{Emo}_\textit{fused}$, expressed as:

\begin{small}
\begin{equation}
\label{eq:fuse
}
\textit{Emo}_{\textit{fused}} = \texttt{CrossAttention}([\textit{Emo}_{\textit{pos}}, H], [\textit{Emo}_{\textit{neg}}, H]).
\end{equation}
\end{small}

\subparagraph{Action Definition: $\text{Intent}_\text{Twice}$.}
The action $\textit{Intent}_\textit{Twice}$ involves selecting an empathetic intent from a predetermined set of $\textit{intent}_{refer}$ (Table~\ref{tab:emo_act}), as determined by the policy network $\mathbbm{M}_{p}$. This network comprises two linear layers and returns a probability distribution $\textit{p}_{act}$ over $\textit{intent}_{refer}$. Consequently, an action is sampled in accordance with this distribution. The importance sampling ratio $\frac{\pi(\textit{Intent}_\textit{Twice}|\textit{Emo}_{\textit{fused}})}{\mu(\textit{Intent}_\textit{Twice}|\textit{Emo}_{\textit{fused}})}$ is employed to rectify any discrepancies within the policy.

\subparagraph{Reward Calculation:}
The reward $r$ is calculated based on how well the selected $\text{Intent}_\text{Twice}$ aligns with the user's emotional state $e$, involving two key components: a reward for positive alignment and a penalty for negative alignment, formally:

\begin{small}
\begin{equation}
\label{eq:r}
R(e) = \begin{cases} 
\texttt{sigmoid}(\textit{Emo}_\textit{pos}[i] \cdot \textit{intent}_\textit{refer}) & \textit{if } \textit{is\_pos}(e) \\
\texttt{sigmoid}(\textit{Emo}_\textit{neg}[i] \cdot \textit{intent}_\textit{refer}) & \textit{otherwise}
\end{cases}
.
\end{equation}
\end{small}
Here, $\textit{intent}_\textit{refer}$ is the selected intent's embedding.

\subparagraph{Correction: Intent Adjustment}
Finally, the intent embeddings are updated through a shared-weight layer during \textit{intent twice} to obtain the final intent \textit{intent} with optimizing cross-entropy loss, denoted as $\textit{L}_\textit{intent}$, ensuring consistency and effectiveness in learning and mimicking empathetic intents.

Overal, the loss for $\textit{intent}_\textit{twice}$ mechanism is represented as $\textit{L}_\textit{twice}$ :
\begin{equation}
\label{eq:loss}
\textit{L}_\textit{twice}=\textit{L}_{\textit{kl}_\textit{pos}}+\textit{L}_{\textit{kl}_\textit{neg}}+ \textit{L}_\textit{intent}.
\end{equation}

\subsubsection{Response Decoding} 
Consequently, We generate the final response using the integrated response-emotion context, $\text{Emo}_\text{fused}$. Following \citet{lin2019moel,majumder2020mime,sabour2022cem,zhou2023case}, We apply a transformer decoder, $\text{TRS}_\text{dec}$ with pointer generator network $P_{\text{Gen}}(*)$, where $\textit{Emo}_\textit{fused}$ serves as both the key and value to predict the word distribution $P_{v}$, as detailed below: 

\begin{small}
\begin{align}
P_{w} &= P(R_t \mid E_{R<t}, \textit{Emo}_{\textit{fused}}, C) \nonumber \\
&= P_{\text{Gen}}(\texttt{TRS}_{\texttt{dec}}(E^C(T_{R<t}), \textit{Emo}_{\textit{fused}}).
\label{eq:gen}
\end{align}
\end{small}
where $E_{R<t}$ denotes the embeddings of all prior responses up to time \textit{t}-1, $E^C(T_{R<t})$ indicates the embedding of the target response, and $P_{\texttt{Gen}}(*)$ represents the pointer generator network \cite{see2017get}. $\text{TRS}_{\text{dec}}$ refers to the transformer decoder function.

\subsection{Training}

Lastly, all parameters of ReflectDiffu are trained jointly in an end-to-end manner to optimize the model by integrating all losses $\textit{L}$, employing loss weight averaging  with hyperparameters $\delta,\zeta,\eta$ as follows:
\begin{equation}
\label{eq:loss}
\textit{L}=\delta\text{L}_\text{em}+\zeta\textit{L}_\textit{twice}+\eta\textit{L}_\textit{res}.
\end{equation}

\section{Experiments Settings}
\begin{table*}[h]
\centering
\renewcommand{\arraystretch}{1}
\resizebox{\textwidth}{!}{%
\begin{tabular}{lccccccccccc}
\toprule
\textbf{Models} & \multicolumn{5}{c}{\textbf{Relevance}} & \multicolumn{2}{c}{\textbf{Controllability}} & \multicolumn{3}{c}{\textbf{Informativeness}} \\
\cmidrule(lr){2-6} \cmidrule(lr){7-8} \cmidrule(lr){9-11}
 & \textbf{B-1 $\uparrow$} & \textbf{B-2 $\uparrow$} & \textbf{B-3 $\uparrow$} & \textbf{B-4 $\uparrow$} & \textbf{BART\textsubscript{Score} $\uparrow$} & \textbf{$\textit{Acc}_\textit{emo} \uparrow$}
 & \textbf{$\textit{Acc}_\textit{Intent} \uparrow$} & \textbf{PPL $\downarrow$} & \textbf{D-1 $\uparrow$} & \textbf{D-2 $\uparrow$} \\
\midrule
MTRS & 17.87 & 8.51 & 4.36 & 2.61 & 0.5173 & 32.96 & - & 37.98 & 0.40 & 1.57 \\
MOEL & 18.02 & 8.67 & 4.35 & 2.73 & 0.5166 & 31.02 & - & 36.81 & 0.43 & 1.76 \\
MIME & 19.82 & 8.86 & 4.43 & 2.77 & 0.5182 & 30.26 & - & 36.93 & 0.51 & 1.92 \\
EmpDG & 19.12 & 8.91 & 4.89 & 2.85 & 0.5171 & 32.90 & - & 37.55 & 0.49 & 1.65 \\
KEMP & 17.92 & 8.54 & 4.38 & 2.71 & 0.5232 & 36.40 & - & 36.59 & 0.66 & 2.43 \\
CASE & 19.66 & 8.95 & 4.92 & 2.90 & 0.5336 & 38.96 & - & 35.97 & 0.70 & 2.66 \\
CAB & 20.23 & 9.39 & 4.96 & 3.01 & 0.5392 & 40.52 & - & 35.06 & 0.89 & 2.95 \\
IAMM & 19.51 & 8.74 & 4.86 & 3.32 & 0.5456 & \underline{\textbf{43.72}} & - & 25.94 & 0.88 & 3.05 \\
\midrule
Harnessing (0-shot) & 6.57 & 2.68 & 1.68 & 1.07 & 0.3881 & 24.40 & - & 230.99 & \textbf{1.79} & \textbf{14.72} \\
Qwen2-7B+CoT & \underline{\textbf{23.31}} & 11.21 & 5.20 & 3.45 & 0.5447 & 23.10 & \underline{\textbf{41.61}} & 25.45 & 0.87 & 3.87 \\
llama-3.1-8B+CoT & 23.38 & \textbf{11.29} & \underline{\textbf{5.25}} & \underline{\textbf{3.47}} & \underline{\textbf{0.5480}} & 21.15 & 32.02 & \underline{\textbf{24.92}} & 0.92 & 4.13 \\
\midrule
\textbf{ReflectDiffu} & \textbf{23.59} & \underline{\textbf{11.25}} & \textbf{5.35} & \textbf{3.62} & \textbf{0.5630} & \textbf{48.76} & \textbf{80.32} & \textbf{24.56} & \underline{\textbf{0.98}} & \underline{\textbf{4.35}} \\
w/o \textit{ERA} & 22.59 & 10.66 & 5.02 & 3.28 & 0.5520 & 42.37 & 78.68 & 24.78 & 0.95 & 4.27 \\
w/o \textit{C-Experts} & 23.13 & 11.05 & 5.06 & 3.31 & 0.5619 & 39.44 & 77.44 & 24.82 & 0.91 & 4.03 \\
w/o \textit{Intent twice} & 20.91 & 9.86 & 4.87 & 3.16 & 0.5436 & 44.56 & 66.44 & 29.25 & 0.85 & 3.97 \\
w/o \textit{EMU} & 21.95 & 10.05 & 4.96 & 3.22 & 0.5490 & 48.35 & 79.24 & 27.45 & 0.69 & 2.96 \\
\bottomrule
\end{tabular}
}
\caption{Results of automatic evaluations and ablation study. Metrics include BLEU-1 to BLEU-4 (B-1~B-4), BARTScore for \textit{Relevance}; emotion and intent accuracy ($\textit{Acc}_{\textit{emo}}$, $\textit{Acc}_{\textit{Intent}}$) for \textit{Controllability}; perplexity (PPL) and distinct-n (D-1, D-2) for \textit{Informativeness}.}
\label{tab:result}
\vspace{-1em}
\end{table*}
\subsection{Dataset}
We evaluate our approach, ReflectDiffu, using the \texttt{EMPATHETICDIALOGUES} dataset \cite{rashkin2018empathetic}, which consists of 24,850 open-domain, multi-turn conversations between two interlocutors where the chatbot provides empathetic responses to the user. 32 emotion categories evenly distributed across all dialogues. Utilizing the ChatGLM4 \footnote{\url{https://huggingface.co/THUDM/glm-4-9b-chat-1m}} \cite{glm2024chatglm,kojima2022large,zhong2024rose} to annotate emotional reasoning within the \texttt{EMPATHETICDIALOGUES} dataset. Additionally, we utilize a fine-tuned Hugging Face model, Commonsense-QA-LLM\footnote{\url{https://huggingface.co/rvv-karma/Commonsense-QA-Mistral-7B}} , to reason and annotate the intents. Ultimately, we extend the original dataset with annotations for emotion reasoning, intent prediction and empathetic dialogue, adhering to the predefined 8:1:1 train/validation/test split.

\subsection{Baselines}
In our experiments, we compare ReflectDiffu with both classic and recent state-of-the-art (SOTA) benchmarks including MTRS \cite{rashkin2018empathetic} , MOEL \cite{lin2019moel} , MIME \cite{majumder2020mime} , EmpDG \cite{li2020empdg} , KEMP \cite{li2022knowledge} , CASE \cite{zhou2023case} , CAB \cite{gao2023cab} and IAMM \cite{yang2024iterative}. Additionally, we incorporate a comparative analysis with Harnessing (0-shot prompting) \cite{qian2023harnessing}, as well as QWen2-7B \cite{yang2024qwen2} and Llama-3.1-8B\cite{dubey2024llama}, two prominent generative language models\cite{laskar2024systematic}. More details about baselines are shown in Appendix~\ref{subsec:baseline}.

\subsection{Implement Details}
ReflectDiffu employs 300-dimensional pre-trained GloVe vectors \cite{pennington2014glove} and follows baselines \cite{rashkin2018empathetic, lin2019moel, majumder2020mime, li2020empdg, gao2023cab, zhou2023case} for a fair comparison. It is implemented in \texttt{PyTorch 2.1.2} and trained on two \texttt{NVIDIA GeForce RTX 4090 GPUs} with a batch size of 32 using NoamOpt as the optimizer with learning rate warmup steps of 6000 and a learning rate decay factor of 0.01. The diffusion step is set to 1000 and the model converges after about 16000 iterations with early stopping.

\subsection{Evaluation Metrics}
\paragraph{Automatic Evaluations.} To assess ReflectDiffu's performance, we use automatic evaluation metrics for relevance, controllability, and informativeness, including $\textit{BLEU-n}$, $\textit{BART}_\textit{Score}$, Emotion Accuracy $\textit{Acc}_\textit{emo}$, Intent Accuracy $\textit{Acc}_\textit{intent}$, $\textit{Distinct-1}$, $\textit{Distinct-2}$, and Perplexity $\textit{PPL}$ (see Appendix~\ref{subpa:auto} for details).

\paragraph{Human Evaluation.} For human evaluation, we conduct A/B testing on empathy, relevance, and fluency with three recruited annotators and a supervisory LLM to resolve disagreements (see Appendix~\ref{subpa:human} for details). We compare ReflectDiffu against a selection of widely adopted baselines and recent generation models to ensure consistency and interpretability in human judgments. For clarity and consistency, models relying heavily on memory mechanisms (IAMM) or general-purpose prompting strategies (Harnessing) are excluded from this evaluation.

\section{Results and Discussion}
\paragraph{Automatic Evaluation Results}
As shown in table \ref{tab:result}, our model, ReflectDiffu, outperforms all baseline models and significantly enhances all metrics. Compared with empathy-specific models\cite{rashkin2018empathetic, lin2019moel, majumder2020mime, li2020empdg, li2022knowledge, zhou2023case, gao2023cab}, which mainly explore the connection between emotion states and empathetic contexts but ignore the internal mechanisms of emotional causes, emotions, and intents and only rely on inferred external knowledge, resulting in suboptimal empathetic controllability (low emotion accuracy $Acc_\textit{emo}$), weak similarity and coherence with the empathetic ground truth (indicated by low $\textit{BLEU-n}$ and $\textit{BART}_\textit{Score}$), and a lack of diversity (implied by low $\textit{Distinct-1}$ and $\textit{Distinct-2}$). In contrast, ReflectDiffu exhibits remarkable superiority, exceeding the best baseline, CAB, approximately in $\textit{BLEU-1}$, $\textit{BLEU-2}$, $\textit{BLEU-3}$, $\textit{BLEU-4}$, $\textit{BART}_\textit{Score}$, $\textit{Acc}_\textit{emo}$ by \textbf{16.6\%} , \textbf{20\%}, \textbf{8.1\%}, \textbf{20.3\%}, \textbf{4.6\%} and \textbf{20.3\%} respectively for Emotion-Contagion Encoder to enhance semantic understanding and achieve \textbf{80.32\%} intent accuracy for its \textit{intent twice} mechanism. Moreover, ReflectDiffu shows improvements of approximately 30.1\% in $PPL$, \textbf{10.1\%} in $\textit{Distinct-2}$, and \textbf{47.4\%} in $\textit{Distinct-2}$ compared with CAB for Diffusion within intent guidance. Compared with Harnessing (0-shot), which performs poorly across relevance, controllability, and fluency (e.g., BLEU-1: 6.57, Acc\textsubscript{emo}: 24.40, PPL: 230.99), ReflectDiffu achieves substantially higher scores while maintaining coherence and diversity, demonstrating its robustness in zero-shot empathetic dialogue. Compared to IAMM, which excels in emotion accuracy (43.72) and diversity (Distinct-2: 3.05) but lacks intent controllability and suffers from higher perplexity, ReflectDiffu achieves superior balance across all dimensions, with higher BLEU scores, better controllability (Acc\textsubscript{intent}: 80.32), and lower PPL. These results highlight ReflectDiffu’s effectiveness in generating empathetic, coherent, and diverse responses.

Moreover, compared with llama-3.1-8B\cite{dubey2024llama} with Chain-of-Thought(CoT) via fewshots (a SOTA LLM-based empathetic dialogue model in our experiments), ReflectDiffu outperforms llama-3.1-8B obviously by 1.90\%,4.32\%,2.73\%,\textbf{130.54\%},\textbf{150.94\%},6.52\% and 5.32\% in $\textit{BLEU-3}$, $\textit{BLEU-4}$, $\textit{BART}_\textit{Score}$, $\textit{Acc}_\textit{emo}$, $\textit{Acc}_\textit{intent}$, $\textit{Distinct-1}$ and $\textit{Distinct-2}$.Higher $\textit{BART}_\textit{Score}$, $\textit{Acc}_\textit{emo}$ and $\textit{Acc}_\textit{intent}$ robustly underscore the efficacy of ReflectDiffu in fostering empathy. Lower $PPL$ and higher $\textit{Distinct-1}$ and $\textit{Distinct-2}$ further corroborate the empathetic diversity that ReflectDiffu can engender.

\paragraph{Human Evaluation Results}Table \ref{tab:human-eval} presents the results of the human A/B testing, comparing ReflectDiffu with various baseline models across three criteria: empathy (Emp.), relevance (Rel.), and fluency (Flu.). The evaluations reveal that ReflectDiffu consistently outperforms the baseline models across all criteria.
\begin{table}[ht!]
\centering
\resizebox{\linewidth}{!}{%
\begin{tabular}{ccccccc}
\toprule
\textbf{Comparison} & \textbf{Aspects} & \textbf{Win} & \textbf{Lose} & \textbf{Tie} \\ 
\midrule
\multirow{3}{*}{ReflectDiffu vs. MTRS} & Emp. & \textbf{51.1} & 18.0 & 30.9 \\ 
 & Rel. & \textbf{48.1} & 17.5 & 34.4 \\ 
 & Flu. & \textbf{40.1} & 11.7 & 48.2 \\ 
\midrule
\multirow{3}{*}{ReflectDiffu vs. MOEL} & Emp. & \textbf{45.4} & 21.2 & 33.4 \\ 
 & Rel. & \textbf{37.3} & 22.5 & 40.2 \\ 
 & Flu. & \textbf{31.4} & 13.7 & 54.9 \\ 
\midrule
\multirow{3}{*}{ReflectDiffu vs. MIME} & Emp. & \textbf{50.3} & 20.8 & 28.9 \\ 
 & Rel. & \textbf{43.7} & 19.2 & 37.1 \\ 
 & Flu. & \textbf{38.4} & 9.1 & 52.5 \\ 
\midrule
\multirow{3}{*}{ReflectDiffu vs. EmpDG} & Emp. & \textbf{52.2} & 19.8 & 27.9 \\ 
 & Rel. & \textbf{50.8} & 16.5 & 32.7 & \\ 
 & Flu. & \textbf{36.4} & 10.3 & 53.3 & \\ 
\midrule
\multirow{3}{*}{ReflectDiffu vs. KEMP} & Emp. & \textbf{55.2} & 23.1 & 21.7 \\ 
 & Rel. & \textbf{62.4} & 29.8 & 7.8 \\ 
 & Flu. & \textbf{35.7} & 13.3 & 51.0 \\ 
\midrule
\multirow{3}{*}{ReflectDiffu vs. CAB} & Emp. & \textbf{53.6} & 22.4 & 24.0 \\ 
 & Rel. & \textbf{56.1} & 24.6 & 19.3 \\ 
 & Flu. & \textbf{32.3} & 10.2 & 57.5 \\ 
\midrule
\multirow{3}{*}{ReflectDiffu vs. CASE} & Emp. & \textbf{52.0} & 15.0 & 33.0 \\ 
 & Rel. & \textbf{45.5} & 25.0 & 29.5 \\ 
 & Flu. & \textbf{49.0} & 27.1 & 23.9 \\ 
\midrule
\multirow{3}{*}{ReflectDiffu vs. Qwen2-7B+CoT} & Emp. & \textbf{52.5} & 22.2 &  25.3 \\ 
 & Rel. & \textbf{53.1} & 25.3 & 21.6 \\ 
 & Flu. & \textbf{41.2} & 12.5 & 46.3 \\ 
 \midrule
 \multirow{3}{*}{ReflectDiffu vs. llama-3.1-8B+CoT} & Emp. & \textbf{51.2} & 21.8 & 27.0 \\ 
 & Rel. & \textbf{54.4} & 24.5 & 21.1 \\ 
 & Flu. & \textbf{33.8} & 18.5 & 47.7 \\ 
\bottomrule
\end{tabular}%
}

\caption{Human A/B evaluation results between ReflectDiffu and baselines.}
\vspace{-1em}
\label{tab:human-eval}
\end{table}

\paragraph{Ablation Study.} As shown in Table \ref{tab:result}, we conducted four ablation studies to evaluate the key components of our model: (1) \textbf{w/o \textit{ERA}}: Removing the Emotion Reason Annotator (\textit{ERA}) that improves emotion understanding with reasoning masks; (2) \textbf{w/o \textit{C-Experts}}: Excluding the Contrastive-Experts for emotion classification; (3) \textbf{w/o \textit{Intent twice}}: Eliminating the Intent \textit{Exploring-Sampling-Correcting} mechanism; and (4) \textbf{w/o \textit{EMU}}: Lacking the Emotion Mimicry Unit (\textit{EMU}) with DDPMs for state representation. 

\subparagraph{Effect of \textit{ERA}.} Excluding Emotion Reason Annotator(\textit{ERA}) designed to improve emotion understanding by reasoning masks leads to a significant decrease in $\textit{BLEU-n}$, $\textit{BART}_\textit{Score}$ and $\textit{Acc}_\textit{emo}$, indicating \textbf{w/o \textit{ERA}} compromises emotion perception and thereby results in inferior empathetic responses' relevance and quality. 

\subparagraph{Effect of \textit{C-Experts}.} Removing Contrastive-Experts \textit{C-Experts} leads to a notable decline in $\textit{Acc}_\textit{emo}$ from 48.76 to 39.44, indicating that \textbf{w/o \textit{C-Experts}} deteriorates the ability to classify emotions, consequently negatively affecting the controllability of empathy, making it harder to precisely match responses with desired emotional states.

\subparagraph{Effect of \textit{Intent twice}.} Eliminating the \textit{Intent Exploring-Sampling-Correcting} mechanism significantly reduced $\textit{Acc}_\textit{intent}$ from 80.32 to 66.44, along with poor $\textit{BLEU-n}$ and $\textit{BART}_\textit{Score}$, higher $\textit{PPL}$, \textbf{w/o \textit{Intent twice}} impairs the model's ability to accurately capture and fulfill response intent, weakening empathetic responses' relevance and quality.

\subparagraph{Effect of \textit{EMU}.} Lacking the Emotion Mimicry Unit (\textit{EMU}) for state representation results in a considerable decrease in $\textit{BLEU-n}$, $\textit{Distinct-1}$ and $\textit{Distinct-2}$, along with $\textit{PPL}$, indicating that \textbf{w/o \textit{EMU}} negatively affects the quality and distinctivenes of empathetic responses.
\begin{table}[!t]
    \centering
    \footnotesize
    \renewcommand{\arraystretch}{0.99}
    \begin{tabularx}{\columnwidth}{p {0.325\columnwidth}|p {0.61\columnwidth}}
        \toprule
        \textbf{Emotion} & Terrified \\
        \midrule
        \multirow{4}{*}{\textbf{Context}}& Yeah about 10 years ago I had a horrifying experience. It was 100\% their fault but they hit the water barrels and survived. They had no injuries but they almost ran me off the road. \\
        \midrule
        \textbf{MTRS} & that is pretty scary ! i am glad you are ok .\\
        \textbf{MOEL} & that is so terrible! i am so sorry. \\
        \textbf{MIME} & oh no ! i am so sorry to hear that .  \\
        \textbf{EmpDG} & oh no , i am so sorry to hear that . \\
        \textbf{KEMP} & oh no ! i hope you are okay .\\
        \textbf{CASE} & i hope you can get it fixed. Are you okay now? \\
        \textbf{CAB} & I hope you are able to get it fixed,and hope you are ok! \\
        \textbf{IAMM} & That sounds really scary and I'm glad you're okay. \\
        \midrule
        \textbf{Harnessing} & So scary! Glad you're okay—stay safe out there. \\
        \textbf{Qwen2-7B+CoT} & I'm sorry to hear about your experience. It sounds stressful and dangerous.\\
        \textbf{llama3.1-8B+CoT} & I'm sorry to hear that. If you want, you can talk more about it. \\
        \midrule
               
        \multirow{3}{*}{\textbf{ReflectDiffu}} & 
        $\textbf{\textit{Intent}}_\textbf{\textit{first}}$: \textit{encouraging}{\color{red}$\times$}\\
        & $\textbf{\textit{Intent}}_\textbf{\textit{twice}}$:\textit{consoling}{\color{green}$\checkmark$} \\
        & oh no! That sounds absolutely terrifying . I hope you were not hurt, Were you injured ? \\
        \midrule
        \multirow{1}{*}{\textbf{Golden}} & Did you suffer any injuries?  \\
        \bottomrule
    \end{tabularx}
    \caption{Case study comparison between ReflectDiffu and baselines.}
    \vspace{-1.4em}
    \label{tab:comparison}
\end{table}
\paragraph{Case Study.}

In this case (Table~\ref{tab:comparison}), ReflectDiffu shows improved empathetic response generation by identifying and mimicking the user's emotional state. Using the \textit{Intent Exploring-Sampling-Correcting} mechanism, the model refines its initial intent from \textit{encouraging} to \textit{consoling}, resulting in a more supportive reply. Compared to baselines, ReflectDiffu better aligns with users' emotions, offers a clear and empathetic follow-up, enhancing interaction quality. (Details on mitigating emotion recognition errors are  provided in Appendix~\ref{subsec:appendx-case}.)

\section{Conclusion}
In this paper, we propose ReflectDiffu, a novel psychological multi-task framework for empathetic dialogue that integrates Emotion-Contagion Encoder and Response Generation Decoder guided by an \textit{Intent Twice} mechanism to better understand users’ emotional states, predict intents accurately, and generate highly intent-aligned empathetic responses. Both automated and human evaluations demonstrate that ReflectDiffu excels in relevance, controllability, and informativeness of empathetic dialogue. Our research may inspire future studies on modeling emotion-intent interaction in human discourse and other linguistic behaviors.

\section*{Limitations}
Our ReflectDiffu framework, integrating emotion contagion and intent prediction mechanisms with the \textit{Intent Twice} mechanism, has performed exceptionally in both automatic and human evaluations, significantly enhancing the relevance, controllability, and informativeness of empathetic responses.

We discuss the primary limitation of this work as follows: The integration of Denoising Diffusion Probabilistic Models (DDPMs) and reinforcement learning mechanisms has augmented the computational requirements for training, presenting challenges for deployment in resource-constrained settings or on devices with limited capabilities. To alleviate this limitation, we have adopted reparameterization and multi-task techniques for optimization. As a result, the overall training time is notably shorter than that of multi-stage LLM \cite{chen2024cause,yang2024qwen2,dubey2024llama} while achieving state-of-the-art outcomes.

In conclusion, despite the existing limitations, ReflectDiffu is relatively lightweight compared to LLM. Moreover, our ongoing research efforts aim to achieve lightweight quantization to accelerate the model's implementation and collaboration.

\section*{Ethical Considerations}
Our research utilizes the \texttt{EMPATHETICDIALOGUES} dataset \citet{rashkin2018empathetic}, an open-source resource devoid of any personal privacy information. To annotate the data for emotion reasoning and intent prediction, we leverage prompts teqhniques \cite{kojima2022large} and LLM contrastive voting mechanisms \cite{zhong2024rose} to label intent and emotional reason, thereby minimizing human bias and reducing the risk of model hallucination. Our human evaluations are conducted by three professional annotators, who operate anonymously to protect privacy and ensure objective assessments following our instructions (refer to Appendix~\ref{subsec:instruction}). Annotators are compensated fairly for their contributions. 

\bibliography{custom}

\appendix

\section{Experimental Details}
\subsection{Baselines}
\label{subsec:baseline}
In our experiments, we compare ReflectDiffu with both classic and recent state-of-the-art (SOTA) benchmarks.

\begin{itemize}[itemsep=0em, parsep=0pt, topsep=0pt]
    \item \textbf{Multitask-Transformer(MTRS):}\citet{rashkin2018empathetic} introduced a Transformer model trained for both sentiment detection and empathetic response generation.
    \item \textbf{MOEL:} \citet{lin2019moel} proposed a Transformer model with 32 emotion-specific decoders and a meta-listener to generate contextually appropriate responses.
    \item \textbf{MIME:} \citet{majumder2020mime} combined a Transformer with a VAE to generate empathetic responses by mimicking user emotions through polarity-based clustering and stochastic emotion mixtures.
    \item \textbf{EmpDG:} \citet{li2020empdg} used a Transformer with a WGAN to capture emotional nuances via a token-level perception mechanism.
    \item \textbf{KEMP:} \citet{li2022knowledge} proposed leveraging external knowledge, including commonsense and emotional lexical knowledge, to enhance empathetic dialogue generation.
    \item \textbf{CASE:} \citet{zhou2023case} integrated a commonsense cognition graph and an emotional concept graph to align user cognition and affection for empathetic responses.
    \item \textbf{CAB:} \citet{gao2023cab} integrated cognition, affection, and behavior to enhance empathetic dialogue generation.
    \item \textbf{IAMM:} \citet{yang2024iterative} improves empathetic response quality by modeling internal affect memory and multi-level affective matching. It is designed to enhance emotional alignment and content diversity.
    \item \textbf{Harnessing (0-shot):} \citet{qian2023harnessing} leverages GPT-4o to generate empathetic responses via zero-shot setting under 30 budget tokens.
    \item \textbf{QWen2-7B + CoT:}
    We fine-tune QWen2-7B\citet{yang2024qwen2}, and then employ Chain-of-Thought (CoT) to infer emotion, intent, and generate empathetic responses for improved empathy.
    \item \textbf{llama3.1-8B + CoT:}
    We fine-tune llama3.1-8B\citet{dubey2024llama}, and then employ Chain-of-Thought (CoT) to infer emotion, intent, and generate empathetic responses for improved empathy.
\end{itemize}

\subsection{Evalutions Metrics}
\paragraph{Automatic Evaluation.}
\label{subpa:auto}To assess ReflectDiffu's performance, we use automatic evaluation metrics in three categories: relevance, controllability, and informativeness. 

\begin{itemize}[itemsep=0em, parsep=0pt, topsep=0pt]
    \item \textbf{Relevance}: We use \textit{BLEU} \cite{papineni2002bleu} and $\textit{BART}_\textit{Score}$\cite{yuan2021bartscore} to measure similarity between generated and reference texts. Higher scores indicate more relevant outputs.
    
    \item \textbf{Controllability}: This is measured by Emotion Accuracy ($Acc_{emo}$) and Intent Accuracy ($Acc_{intent}$), which check the model's ability to detect emotions and recognize user intent.
    
    \item \textbf{Informativeness}: Evaluated using Distinct-1, Distinct-2 \cite{li2016diversity}, and Perplexity (PPL) \cite{serban2015hierarchical}. 
    \begin{itemize}
        \item \textbf{Distinct-N}: Measures the proportion of unique unigrams and bigrams, indicating diversity. Higher scores show more varied responses.
        \item \textbf{Perplexity (PPL)}: Lower PPL scores indicate better performance, as the model predicts the next word more accurately, resulting in more fluent and coherent text.
    \end{itemize}
\end{itemize}

\paragraph{Human Evaluation.}
\label{subpa:human}Following \citet{zhou2023case,gao2023cab,wang2024ctsm}, we conduct Human A/B testing between response pairs based on the following criteria evaluated by three annotators: \textbf{(1) Empathy (Emp.):} Assessing the model's ability to generate empathetic content, including understanding the user's emotional state and responding appropriately. \textbf{(2) Relevance (Rel.):} Determining how well the model's responses relate to the dialog history, ensuring coherence and logical progression. \textbf{(3) Fluency (Flu.):} Evaluating the naturalness and readability of the replies, including grammatical correctness and ease of understanding. To ensure fair scoring, we introduced a supervisory LLM, ChatGPT\footnote{\url{https://chat.openai.com}} inspired by \citet{zheng2024judging}. In cases of significant disagreement among annotators, ChatGPT provided the final rating.




\section{Additional Experiments}
\subsection{Explanation of the hyperparameter $n$ of $\text{intent}_\text{infer}$}
\label{subsec:Explanation_of_hyperparameter_n_of_intent_infer}
\begin{figure}[ht]
  \includegraphics[width=\linewidth]{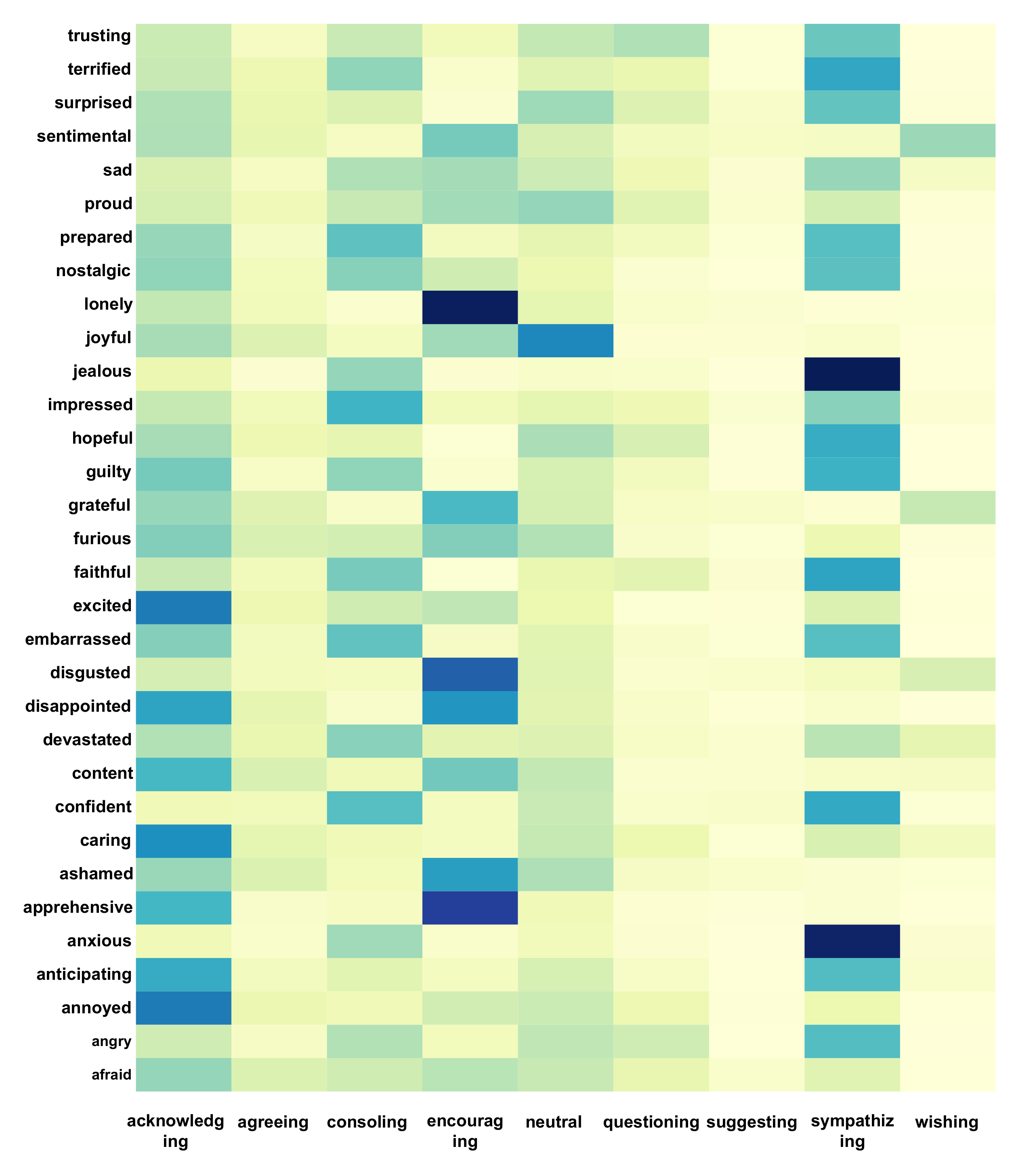}
  \caption{Heatmap of Relative Frequencies of Empathetic Intentions for Each Emotion.}
  \label{fig:freheat}
\end{figure}

After annotating the dataset with empathetic intentions, we conducted a statistical analysis to determine the frequency of each intention for every emotion. Figure~\ref{fig:freheat} illustrates this data, where rows represent distinct emotions and columns represent specific empathetic intentions. The color intensity in each cell indicates the relative frequency of a particular intention corresponding to an emotion, with darker shades signifying higher frequencies. Figure~\ref{fig:freheat} aids in understanding the predominant empathetic actions associated with each emotional state, thereby providing insights into the alignment of universal intents ($\text{Intent}_{\text{refer}}$) with user emotions. We observed that setting $n=3$ effectively avoids non-universal intentions while ensuring that, besides the neutral intent, a more meaningful intent is sampled within the top-2 $\text{Intent}_{\text{refer}}$.

\label{subsec:appendx-case}
\begin{table}[!t]
    \centering
    \footnotesize
    \renewcommand{\arraystretch}{0.99}
    \begin{tabularx}{\columnwidth}{p {0.33\columnwidth}|p {0.6\columnwidth}}
        \toprule
        \textbf{Emotion} & hopeful \\
        \midrule
        \multirow{2}{*}{\textbf{Context}}& i just applied for graduate school ! i feel good about my chances !  \\
        \midrule
        \textbf{MTRS} & i hope you have a great time ! \\
        \textbf{MOEL} & i am sure you will do great . \\
        \textbf{MIME} & i am sure you will do great !   \\
        \textbf{EmpDG} & that is great ! i hope you are going to school for a new one ? \\
        \textbf{KEMP} & that is great ! i hope you get it .\\
        \textbf{CASE} & That is good, I am glad you are able to get it \\
        \textbf{CAB} & That is awesome! glad you are better ! \\
        \textbf{IAMM} & Perfect! Graduate school applications are exciting and I hope it works out for you.\\
        \midrule
        \textbf{Harnessing} & That's great! Hope it goes well for you. \\
        \textbf{Qwen2-7B+CoT} & That's wonderful! Applying for graduate school is a significant step\\
        \textbf{llama3.1-8B+CoT} & Congratulations on taking this important step! That's fantastic! \\
        \midrule
               
        \multirow{3}{*}{\textbf{ReflectDiffu}} & 
        $\textbf{\textit{emotion:}}$ joyful {\color{red}$\times$ }\\
        & $\textbf{\textit{Intent}}_\textbf{\textit{first}}:$  acknowledging{\color{red}$\times$}\\
        & $\textbf{\textit{Intent}}_\textbf{\textit{twice}}$:consoling{\color{green}$\checkmark$} \\
        & i am proud and sure you'll do just fine in school. \\
        \midrule
        \multirow{1}{*}{\textbf{Golden}} & I'm so proud of you! I'll pray for your success!  \\
        \bottomrule
    \end{tabularx}
    \caption{Case study in misclassification comparison between ReflectDiffu and baseline models.}
    \vspace{-1.3em}
    \label{tab:comparison2}
\end{table}

\subsection{Case Study in Misclassification}

We deliberately selected a ReflectDiffu emotion recognition error case to validate the effectiveness of our reflection mechanism. Table~\ref{tab:comparison2} compares responses from various models, including MOEL, MIME, EmpDG, KEMP, CASE, CAB, IAMM, Harnessing, Qwen2-7B+CoT, llama3.1-8B+CoT and ReflectDiffu, to a user's context of feeling hopeful after applying for graduate school. Initially, ReflectDiffu misidentified the emotion as "joyful" and the intent as "acknowledging." However, after employing the reflection mechanism, it correctly identified the intent as "encouraging." This demonstrates the model's capability to correct errors and generate more empathetic responses through its reflection mechanism.

\section{Implement Details}
\subsection{Emotion Reason Annotator}
\label{subsec:appendix-era}
Our approach leverages BERT \cite{devlin2019bert}, an attention-based semantic composition network, and conditional random fields (CRF) to effectively annotate emotional phrases with tags such as \texttt{<$em$>} or \texttt{<$noem$>}.
Specifically, given the token-level dialogue history \( C \), where \( w_i^j \) represents the \( i \)-th token in the \( j \)-th utterance, we use a pretrained BERT model to obtain contextualized token representations \( h^i_j \):
\begin{equation}
  \label{eq:bert}
  {h}^i_j = \texttt{BERT}(w^i_j).
\end{equation}
where \( h^i_j \) is the hidden state output by \texttt{BERT} corresponding to the token \( w_i^j \).

Unlike traditional Named Entity Recognition (NER) models \cite{souza2019portuguese,qi2023ssmi}, we introduce an attention-based semantic composition network that progressively distinguishes between binary sets of words.

Each token representation \( h^i_j \) is initially treated as a word-level feature representation. The attention network computes the correlation between pairs of word vectors \( h^i_j \) and \( h^k_m \). The relevance score \( \alpha_{ik}^{jm} \) and reasoning representation \( \tilde{h}\) are defined as:
\begin{small}
\begin{equation}
\label{eq:attention}
\alpha_{ik}^{jm} = \frac{\exp\left(\texttt{Attention}(h^i_j, h^k_m)\right)}{\sum_{k,m} \exp\left(\texttt{Attention}(h^i_j, h^k_m)\right)},
\end{equation}
\end{small}
\begin{small}
\begin{equation}
\label{eq:reason}
\tilde{h}^i_j = \sum_{k,m} \alpha_{ik}^{jm} h^k_m.
\end{equation}
\end{small}
where \(\tilde{h}^i_j\) is the attention-weighted representation for the token \( w_i^j \), enriched with contextual information from related tokens within the conversational turn.

Finally, the enriched representations are passed through a CRF layer to obtain the final predictions:

\begin{small}
\begin{equation}
\label{eq:crf}
P(\mathbf{r}\mid\mathbf{\tilde{h}}) = \frac{\exp\left(\sum_{j=1}^{n} \left( A_{r_{j-1}, r_j} + \mathbf{W}{r_j} \tilde{h}^i_j \right)\right)}{\sum_{\mathbf{y}' \in \mathcal{R}(\mathbf{\tilde{h}})} \exp\left(\sum_{j=1}^{n} \left( A_{r'_{j-1}, r'_j} + \mathbf{W}{r'_j} \tilde{h}^i_j \right)\right)}.
\end{equation}
\end{small}
where \( \mathbf{r}=(r_1, r_2, \ldots, r_n) \) represents the sequence of reasoning labels, each \( y_i \in \{\texttt{<$em$>}, \texttt{<$noem$>}\} \), \( \tilde{h}^i_j \) is the reasoning representation, \( A \) is the transition matrix, and \( \mathbf{W} \) represents the weights for the CRF layer.
\subsection{Definition of $L_{\text{em}}$}
\label{subsec:appendix-loss}
Inspired by \cite{chen2020simple,zheng2023robust}, we use the NT-Xent loss (\(n_{\text{emo}} = 32\)) \(L_{\text{NTX}}\) and cross-entropy loss (\(L_{\text{ce}}\)) for contrastive emotion classification (\(L_{\text{em}}\)) , formally:

\begin{small}
\begin{gather}
L_{\text{NTX}}^{(n_{\text{emo}})^i} = -\sum_{i=1}^{n} \sum_{\substack{j=1 \\ j \neq i}}^{n} \mathbbm{1}_{[y_i = y_j]} \log \frac{\exp(\text{s}(i, j))}{\sum_{k=1}^{n} \exp(\text{s}(i, k))}, \\
L_{\text{NTX}} = \sum_{i=1}^{32} L_{\text{NTX}}^{(n_{\text{emo}})^i}, \\
L_{\text{cls}} = -\log \mathcal{P}[e],\\
L_{\text{em}}  = L_{\text{NTX}} + L_{\text{cls}}.
\end{gather}
\end{small}

Here, $n$ represents the number of samples, $y_i$ is the pseudo-label of the $i$-th sample, $\mathbbm{1}_{[y_i = y_j]}$ is an indicator function that equals 1 if $y_i = y_j$ and 0 otherwise, $\text{s}(i, j)$ denotes the similarity between samples $i$ and $j$, and $\mathcal{P}[e]$ is the predicted probability for the true emotion class $e$.

\section{Annotators Instructions for Human Evaluation}
\label{subsec:instruction}
Professional annotators received our detailed guidelines to guarantee high-quality and unbiased evaluations.
\begin{itemize}[itemsep=0em, parsep=0pt, topsep=0pt]
    \item \textbf{Evaluation Criteria}: Annotators assessed responses based on three key criteria:
    \begin{itemize}
        \item \textbf{Empathy (Emp.)}: Evaluators were instructed to assess how well the response understood and mirrored the user's emotional state. Examples of high empathy included responses that acknowledged the user's feelings and provided appropriate support or encouragement. Low empathy responses were those that failed to recognize or appropriately respond to the user's emotions.
        \item \textbf{Relevance (Rel.)}: This criterion focused on how well the response related to the previous conversation context. High relevance responses directly addressed the user's statements or questions, maintaining coherence. Low relevance responses were off-topic or did not logically follow the conversation flow.
        \item \textbf{Fluency (Flu.)}: Evaluators assessed the grammatical correctness and naturalness of the responses. Fluent responses were well-structured, easy to read, and free of grammatical errors. Non-fluent responses contained grammatical mistakes, awkward phrasing, or were difficult to understand.
    \end{itemize}
    \item \textbf{Conflict Resolution}: Procedures were established to handle disagreements among annotators:
    \begin{itemize}
        \item When annotators disagreed on the evaluation of a response, a discussion was initiated to reach a consensus.
        \item If consensus could not be achieved, a supervisory Large Language Model (LLM) provided the final rating to ensure objective and consistent evaluations across different cases.
    \end{itemize}
    \item \textbf{Anonymity and Privacy}: Annotators were assured that their evaluations would be anonymized to protect their identities. They were informed that their personal information would not be shared or disclosed in any part of the study, ensuring their privacy and confidentiality.
    \item \textbf{Compensation and Acknowledgment}: Annotators were informed about their compensation:
    \begin{itemize}
        \item They were fairly compensated for their time and effort in evaluating the responses.
        \item Their contributions would be acknowledged in the final publication of the study to recognize their important role in the research process.
    \end{itemize}
\end{itemize}

\end{document}